# Hope Speech Detection on Social Media Platforms

Pranjal Aggarwal [¶],Pasupuleti Chandana[†], Jagrut Nemade [¶], Shubham Sharma [¶],Sunil Saumya [¶],Shankar Biradar [¶]

[¶] Department of Computer Science and Engineering, Indian Institute of Information Technology Dharwad
Dharwad, Karnataka – 580009

[†]Department of Electronics and Communication Engineering, Indian Institute of Information Technology Dharwad
Dharwad, Karnataka – 580009

## Abstract

Ever since personal computers became widely available in the consumer market, the amount of harmful content on the internet has significantly expanded. In simple terms, harmful content is anything online which causes a person distress or harm. It may include hate speech, violent content, threats, non-hope speech, etc. The online content must be positive, uplifting and supportive. Over the past few years, many studies have focused on solving this problem through hate speech detection, but very few focused on identifying hope speech. This paper discusses various machine learning approaches to identify a sentence as Hope Speech, Non-Hope Speech, or a Neutral sentence. The dataset used in the study contains English YouTube comments and is released as a part of the shared task "EACL-2021: Hope Speech Detection for Equality, Diversity, and Inclusion". Initially, the dataset obtained from the shared task had three classes: Hope Speech, non-Hope speech, and not in English; however, upon deeper inspection, we discovered that dataset relabelling is required. A group of undergraduates was hired to perform the entire dataset's relabeling task. We experimented with conventional machine learning models (such as Naïve Bayes, logistic regression and support vector machine) and pre-trained models (such as BERT) on relabelled data. According to the experimental results, the relabelled data has achieved a better accuracy for Hope speech identification than the original data set.

## 1. Introduction

Social media has a significant impact on society. In today's scenario, everyone is free to state their opinions which could be both positive and negative. Constructive criticism is always good, but the way things are turning in today's world, people sometimes misuse their freedom of speech on social media by trolling and bullying others. Social media has roots deep in our society, and multiple studies have found a strong link between heavy social media and an increased risk for depression, anxiety, loneliness, self-harm, and even suicidal thoughts[26].

Social media has many merits, including access to opinions on various topics, meeting and getting to know different people, spreading art/culture and promoting talent, giving support and strength to the voiceless, and spreading awareness. However, it also comes with certain demerits like spreading false information, hatred, cyberbullying, and manipulating people's views, to name a few. The unfriendly content of social media has many consequences, such as reducing one's confidence, ruining mental health, degrading moral values, and so on. Thus it is crucial and necessary to remove unfriendly content to fight out all the cons of social media and make it a better platform for everyone.

Over the last decade, the impact of social media on an individual has increased exponentially. People these days express their views on social media on subjects such as women in the fields of Science, Technology, Engineering, and Management (STEM). People belong to the Lesbian, Gay, Bisexual, Transgender, Intersex Queer/Questioning (LGBTQ) community, racial minorities or people with disabilities. Due to this freedom of speech, delicate issues such as discrimination against minorities, criticism, racism etc., have also increased. To preclude this, researchers have come up with various methods such as hate speech detection (Schmidt and Wiegand, 2017)[1], offensive language identification (Zampieri et al., 2019a)[2] and abusive language detection (Lee et al., 2018)[3]. But these techniques for abusive language detection are fallacious as they do not consider the potential biases of the data set. Consequently, this bias in the dataset causes abusive language detection to be inclined towards one group over the other. Therefore, researchers have turned their work to promote hope speech rather than eliminating harmful content.
Previous research has highlighted the importance of identifying unsavoury content on social media platforms and attempted to identify content like hate speech and fake news[5][7][9][10]. Still, it is also important to promote and encourage content that offers support, reassurance, suggestion, and inspiration. This promotion of righteous content would inspire people and prevent the spread of harmful content such as racial/sexual remarks, nationally motivated slurs, and so on.

In the dataset published by hugging face to identify hope speech, the researchers have divided the dataset into two categories (Hope and NonHope). They have considered expressions that offer support, reassurance, suggestions, inspiration, and insight as Hope speech . Expressions that do not bring positivity, such as racial comments, comments on ethnicity, sexual comments, and nationally motivated slurs as Non-Hope Speech. They have trained the dataset on various Machine Learning and Deep Learning models to classify Hope and Non-Hope Speech. However, the problem with this dataset is that it contains various discrepancies detailed in section 4.1. To solve this problem, this paper tries to fix those discrepancies and present the dataset with improvements[27].

The main contribution of this work focuses on:

i. Dealing with anomalies in the original dataset by introducing a new label named "Neutral speech."
ii. Validating the updated dataset on various classifiers like Naïve Bayes, Logistic Regression, SVM and a transformer learning technique such as BERT.
iii. The proposed dataset could pave the path for further research endeavours. This dataset can also be used on platforms like YouTube, Facebook, Instagram, etc., for content moderation.

The rest of the paper is organized in the following manner: Section 2 briefly describes the related works in the area. Section 3 contains information about the dataset and the methods used for pre-processing and classification. Further, section 4 discusses the results obtained, while Section 5 includes a Conclusion and Future Work.

## 2. Related Works

More recently, the identification of hope speech on social media has caught the research community's interest. Previously, most studies on hate speech and fake news identification had been published, but hope speech incorporates all of these into a single spectrum. A hope speech is something that spreads positivity, motivates others, and raises hopes for a better tomorrow(Chakravarthi and Muralidaran, 2021)[4]. The antithesis of this will be a non-hope speech. Because hope speech identification in social media websites is a relatively new topic with very few studies focusing on it, this section will address all studies relating to toxic content identification on social media websites. Several strategies for identifying harmful social media content have been discussed.

### 2.1. Fake news detection

Detecting fake news has been part of various researches in the past. (Castillo et al., 2011)[5] (Biradar et al., 2022)[6] discovered that lexical parameters such as word count, number of different words, and characters can help identify fake content in a twitter data set. According to several types of research, in addition to linguistic features, syntactic features such as the number of keywords, sentiment score, polarity, and POS tagging are also helpful in distinguishing false news from authentic news(Hassan et al., 2010)[7]. Some studies have attempted to distinguish fake news by examining social aspects such as unique features and group attributes, along with individual characteristics such as "age," "sex," and "occupation,". It also includes online user behaviours such as "number of comments" and "number of followers" (Castillo et al., 2011)[5]. People have recently attempted to investigate propagation-based features for detecting fake news. (Shu et al., 2020)[8] created a hierarchical distribution network for fake and legitimate news, and conducted a rigorous evaluation of structural, temporal, and linguistic features to identify fake news.

### 2.2. Hate speech detection

Another category of research focuses on hate speech identification. (Biradar et al., 2021)[9] (Biradar and Saumya, 2022)[10] developed a translation-based approach for detecting objectionable text in Hindi-English coding mixed data. Some researchers have also experimented with language models such as fine-tuned BERT to detect hate news (Mozafari et al.,2019)[11]. (Chopra et al., 2020)[12] Explained how targeted hatred embeddings mixed with Social-Network-based characteristics outperform existing state-of-the-art models. (Santosh and Aravind, 2019)[13] created a sub-word level LSTM and hierarchical LSTM model with attention. (Roy et al., 2020)[22] used Deep Convolutional Neural Network and LSTM network to detect hate speech in Twitter data. Hate speech detection has also been extended to Dravidian languages by (Roy et al., 2022)[21] using transformer-based models like m-BERT, distilBERT, and xlm-Roberta that performed better than the ML and DL-based models.

### 2.3. Hope speech detection

In recent years, hope speech has replaced the aforementioned methodologies. (Chakravarthi,2020)[14] made the first such effort, they created a hope speech data set in multiple languages, including Tamil, Malayalam, and English, as part of a shared task to support the study of positive content identification through social media. (Dowlagar and Mamidi, 2021)[15] used Transfer learning model mBERT combined with CNN to identify

Hope speech in the English data set. Few researchers have focused on various machine learning and deep learning models and their ensemble settings, and ensemble settings have been discovered to provide better results for Hope speech recognition(Saumya and Mishra, 2021)[16]. (Roy et al., 2022)[19] used machine learning techniques like Logistic Regression, Random Forest, Naïve Bayes and Extreme Gradient Boosting. (Kumar et al., 2022)[20] used word level TF-IDF and character level TF-IDF features with an ensemble model to classify Hope and non-hope speech. (Junaida and Ajees,2021)[17] has also worked with context-aware embeddings using RNN models. Finally, in addition to transformer models, a few researchers have focused on linguistic variables such as TF-IDF and Character N-grams, using classic ML algorithms such as L.R. and SVM, to distinguish between Hope and Non-hope Speech (Dave et al., 2021)[18].

### 2.4. Cyberbullying and Fake Profile Detection

Detecting cyberbullying and fake profiles can also be considered harmful content detection. (Roy et al., 2022)[23] has used an ensemble machine learning model to identify cyberbullying through various machine learning classifiers and voting-based ensemble learning. (Roy et al., 2022)[24] has also used machine learning classifiers like Naïve Bayes, Logistic Regression and Random Forest to address the issue of cyberbullying.

Fake accounts across various social media platforms are generally used for specific purposes. One such purpose is to target certain users to deliver harmful content to them. (Roy and Chahar, 2020)[25] have summarized the recent technological advances to detect fake accounts and their challenges and limitations.

## 3. Data and Methods

### 3.1. Dataset Description

The dataset[14] consists of data on recent topics of Equality, Diversity and Inclusion, including women in STEM, LGBTQ, COVID-19, Black Lives Matter, United Kingdom (U.K.) versus China, United States of America (USA) versus China and Australia versus China from YouTube video comments. The original dataset obtained from Hugging Face consists of 3 classes: Hope, non-hope, and Not English having 22762 training data and 2843 testing data as described in Table 1.

| | **Hope** | **Non-Hope** | **Not English** |
|---|---|---|---|
| **Train** | 25940 | 2484 | 27 |
| **Test** | 268 | 2552 | 2 |

Table1: Original Dataset Distribution

After reassigning the labels, we removed the sentences which were not in English (27 in total). We introduced a new label named "Neutral"; the updated data distribution is shown in Table 2.

| | **Hope** | **Non-Hope** | **Neutral** |
|---|---|---|---|
| **Train** | 5419 | 8024 | 7792 |
| **Test** | 706 | 990 | 1072 |

Table 2: Relabelled Dataset Distribution

In this study, we implement methods for classifying Hope speech and Non-Hope speech contents. Several Traditional machine learning, deep learning, and Transfer learning approaches were implemented to accomplish the purpose. The subsequent sections will provide the detail of all the proposed approaches.

## 3.2. Revisiting the dataset

The original dataset was analyzed using different techniques of Exploratory Data Analysis. One of these was to analyze the data using word clouds. It was found that words like Life, Matter, People, etc., were almost the same size for both Hope and Non-Hope labels (meaning a similar number of occurrences in both), as shown in Fig. 1 and Fig. 2 below.

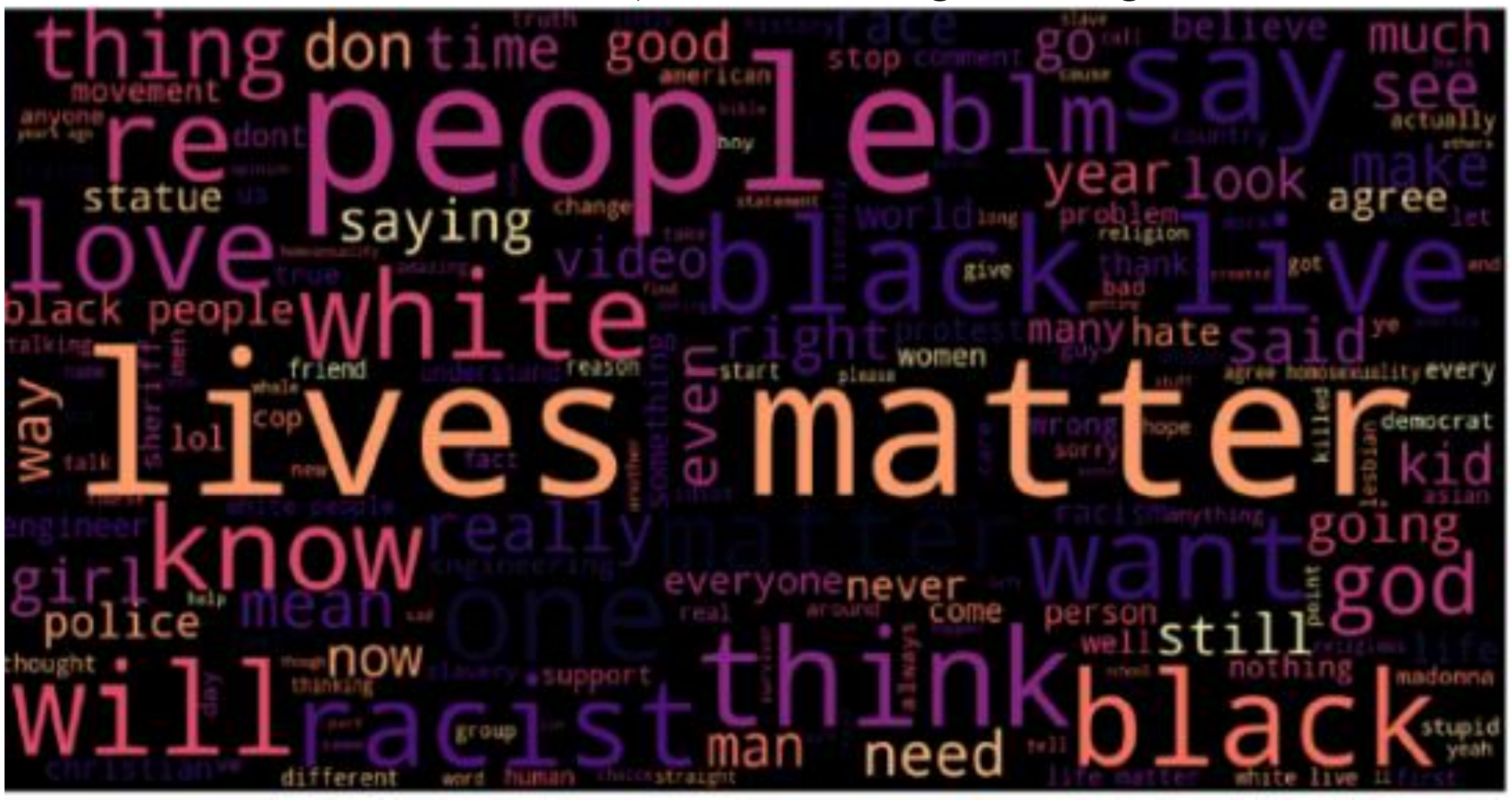

*Figure 1: Word cloud of non-hope class*

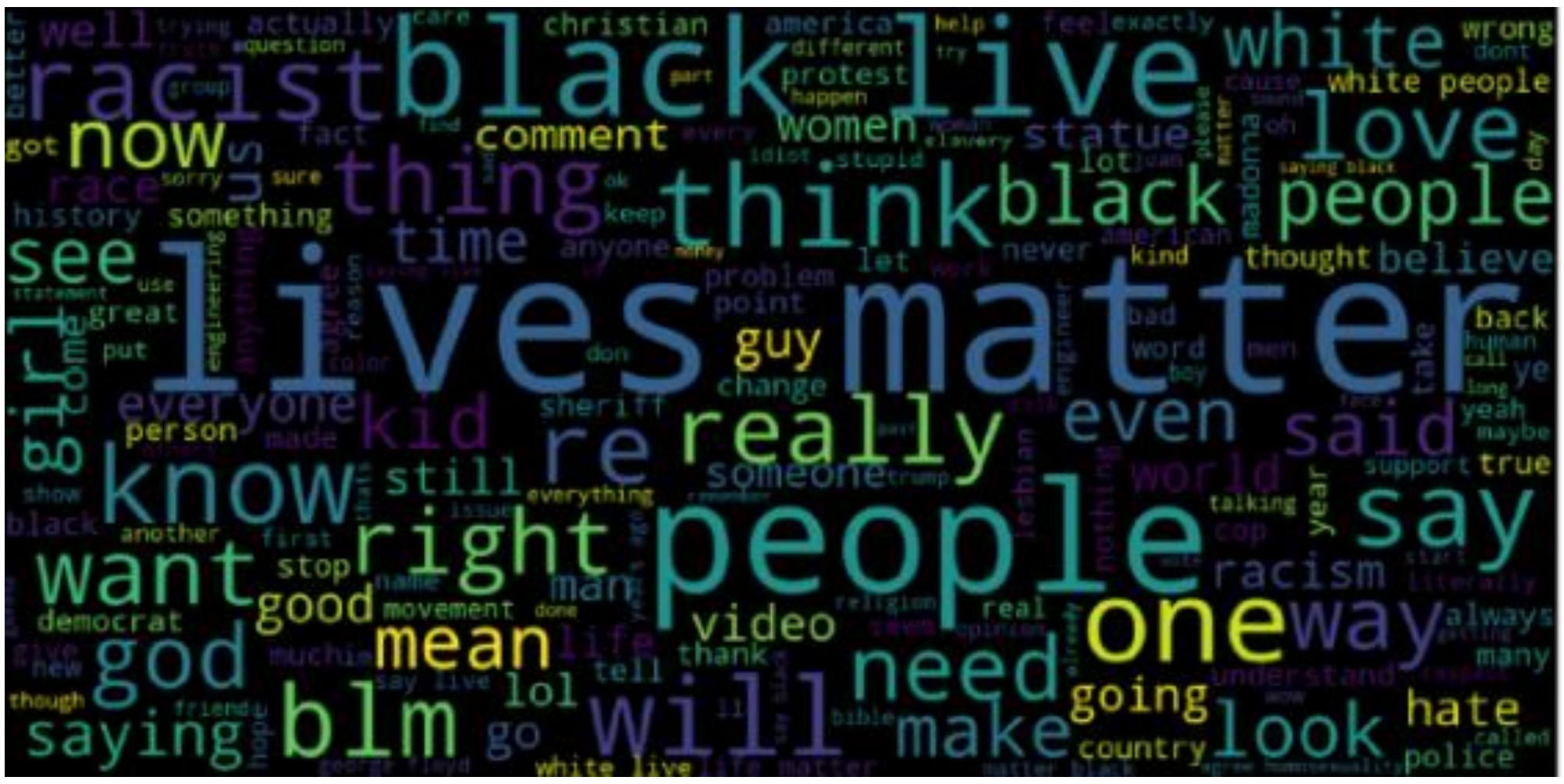

*Figure 2: Word cloud of hope class*

### 3.2.1. Reasons to relabel the dataset

Building up from the Exploratory Data Analysis, we delved deeply into the dataset to find the following vulnerabilities occurring frequently:

1. The same comment has been labelled as Hope and Non-hope at different occurrences, making it difficult for the model to be trained accurately. For example, "Madonna is God" is labelled as both Hope and Non-hope.
2. English comments are labelled as Not-English. For example, "@cubicPas123 it's not the right time" is labelled as Not-English.
3. Hopeful comments are labelled as Non-hope. For example, "God gave us a choice" is labelled as Non-hope.
4. Non-hope comments are labelled as Hope. For example, "Democrats are racist" is labelled as Hope.
5. Not-English comments are labelled as Hope or Non-hope. For example, "@jessica walker lo pas encantada" is labelled as Hope.

#### 3.2.2. Criteria of relabelling

To address these vulnerabilities and to make the dataset more precise, we relabeled the whole data accordingly. Text Data showing inspiration, support, suggestion, well-being, joy, happiness, optimism, faith, and expression of love are labelled under Hope Speech. Further, negativity, abuse, racial and sexual comments, comments towards nationality, hate towards a minority, and prejudices are labelled under Non-hope Speech. Lastly, remaining statements like incomplete statements, One's opinion, statistical data etc., were labelled under Neutral Speech.

1. Dataset is divided into three categories: Hope, Non-hope and Neutral (Not-English were removed in the pre-processing step).
2. If a comment is inflicting neither Hope nor Non-hope, it is labelled as a Neutral comment. For example, "@paul23 you are right" is considered a neutral comment as it does not give Hope or Non-hope to others.
3. Comments without context are considered Neutral. For example, "@walker, this is not necessary".
4. Comments that might be hopeful to some but non-hopeful to others were labelled as Non-hope comment

To make the dataset robust, we took some samples from the updated dataset and asked some volunteers to classify them as Hope, Non-Hope or Neutral. For this, we took help from 4 volunteers from IIIT Dharwad who were good at reading and understanding the English language. A google form was created using JavaScript comprising sample sentences from the original dataset. Each sentence had to be categorized as Hope, Non-Hope or a Neutral sentence. Upon receiving the responses, the mode (a measure of central tendency) was calculated to finally give a label to all the sentences.

### 3.3. Data Pre-Processing

The dataset was cleaned up and pre-processed before model implementation. The following steps were employed for the same:

1. Unwanted text such as HTML decoding, @mention, URL links, punctuation, etc. were removed
2. All the words were converted to lowercase. This was done because the words "Everything", "everything", and "EVERYTHING" are treated as different words, although they mean the same.
3. All the abbreviations and short forms of words were expanded using contraction mapping. For example, BLM was expanded to black lives matter, coz to because and so on. This was also done to ensure words with same meaning are treated as same words
4. Tokenization is applied to this dataset where the sentences are broken down into tokens and stop words are removed.
5. These tokens are lemmatized by converting the words to their root words. For example total, totally, and totalized were converted into the total.

Apart from the regular contractions used in English (such as isn't, don't, haven't, we've, etc), we found many more contractions that are generally used on social platforms. Table 3 shows some of the many contractions discovered.

| **Contraction** | **Expanded Form** | **Contraction** | **Expanded Form** |
|---|---|---|---|
| blm | black lives matter | ppl | people |
| omg | oh my god | poc | people of color |
| libs | liberals | pov | point of view |
| smh | stupid minded humans | some1 | someone |
| nuf | enough | yt | youtube |
| coz | because | m8 | mate |
| cuz | because | dems | democrats |
| cos | Because | msm | main stream media |

Table 3: Some of the Self Discovered Contractions

In addition to these, there are 99 more contractions which were found during relabelling of the dataset.

To vectorize the statements, we used TF-IDF vectorization. TF-IDF stands for term-frequency inverse document frequency. It normalizes the number of times a particular term appears in the document by dividing the word count in a sentence by the times that word has appeared throughout the dataset.

$$tf(t,d) = \frac{f_d(t)}{max_{wEd} f_d(w)}$$

$$idf(t,D) = \ln \frac{|D|}{|\{d \in D: t \in d\}|}$$

$$tfidf(t,d,D) = tf(t,d) \cdot idf(t,D)$$

$$tfidf'(t,d,D) = \frac{idf(t,D)}{|D|} + tfidf(t,d,D)$$

$$f_d(t) := frequency\ of\ term\ in\ document$$

$$D := corpus\ of\ documents$$

## 3.4. Model Description

The model consists of three units, the pre-processing unit, the feature extraction and data balancing unit, and the classification unit. In the pre-processing unit, the dataset is cleaned up by removing unwanted text portions, expanding the abbreviations, and using lemmatization to convert words to their root words. The output of this step is then fed to the feature extraction unit that uses the TF-IDF vectorization technique to identify the importance of specific words or phrases in the dataset. Therefore, the sentences in the dataset are converted to vectors of different values. These vectors are then used to balance the dataset using data balancing methods like SMOTE and ADASYN by adding some synthetic samples. The output from this unit is given to the final unit, which constitutes various traditional Machine Learning models like Naïve Bayes, Logistic Regression and SVM. These models were implemented using the scikit-learn library available in python. Through various experimental trials, we arrived at some hyperparameters for these models that are shown in table 4.

| Model | Hyperparameters |
|---|---|
| Naïve Bayes | alpha = 1.0 |
| Logistic Regression | penalty = 'l2', C = 1.0, max_iter = 100 |
| SVM | C = 1.0, kernel = 'linear', degree = 3, gamma = 'auto' |

Table 4: Hyperparameters for Traditional ML Models

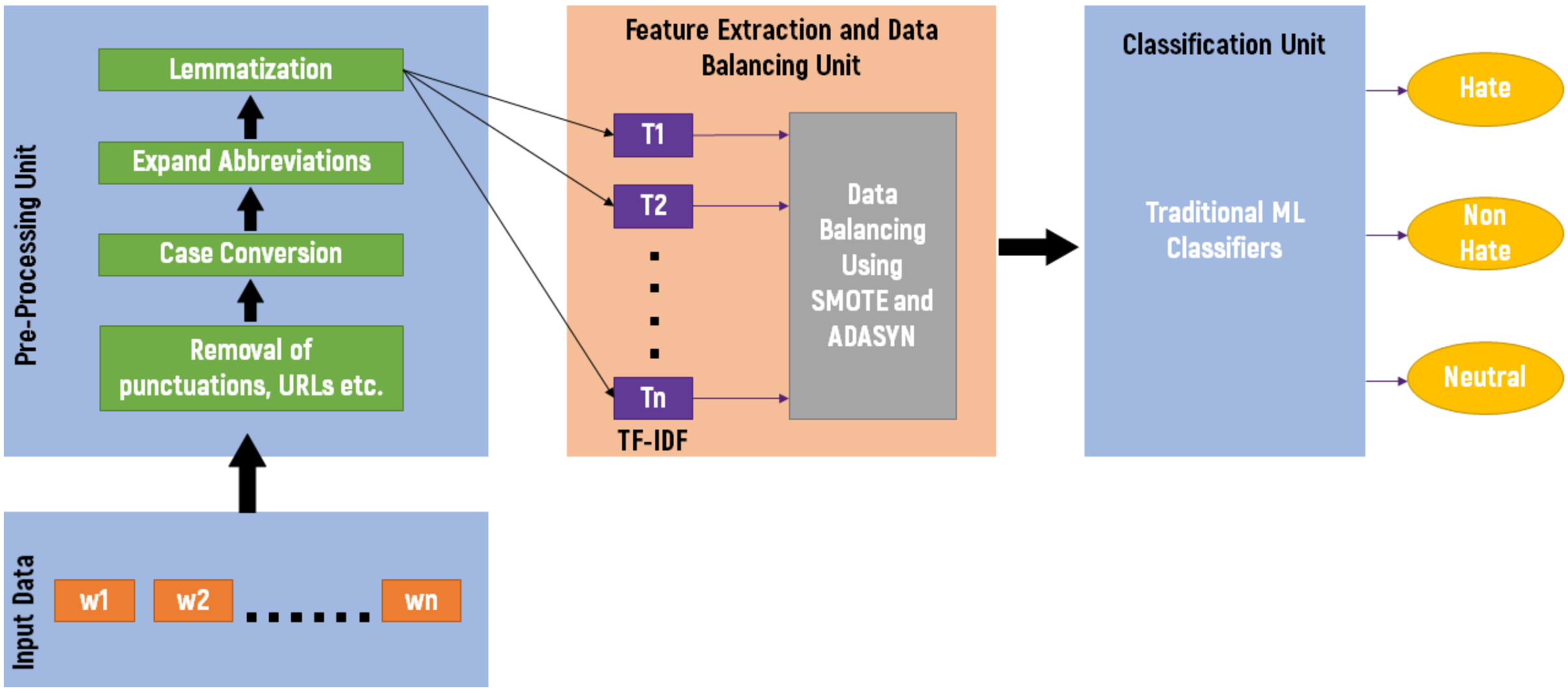

*Figure 3: Traditional ML Models*

One of the reasons we decided to use BERT is that BERT is a non-directional model built on transformers. This means that, unlike other directional models that read the text in a single direction, BERT examines the entire text at once, allowing it to understand the context by considering surrounding words.

After pre-processing the dataset, we used the BERT tokenizer to convert the word's into numerical data. We also added the [CLS] token and [SEP] (shown in Fig. 5) token to implement the next sentence prediction strategy.

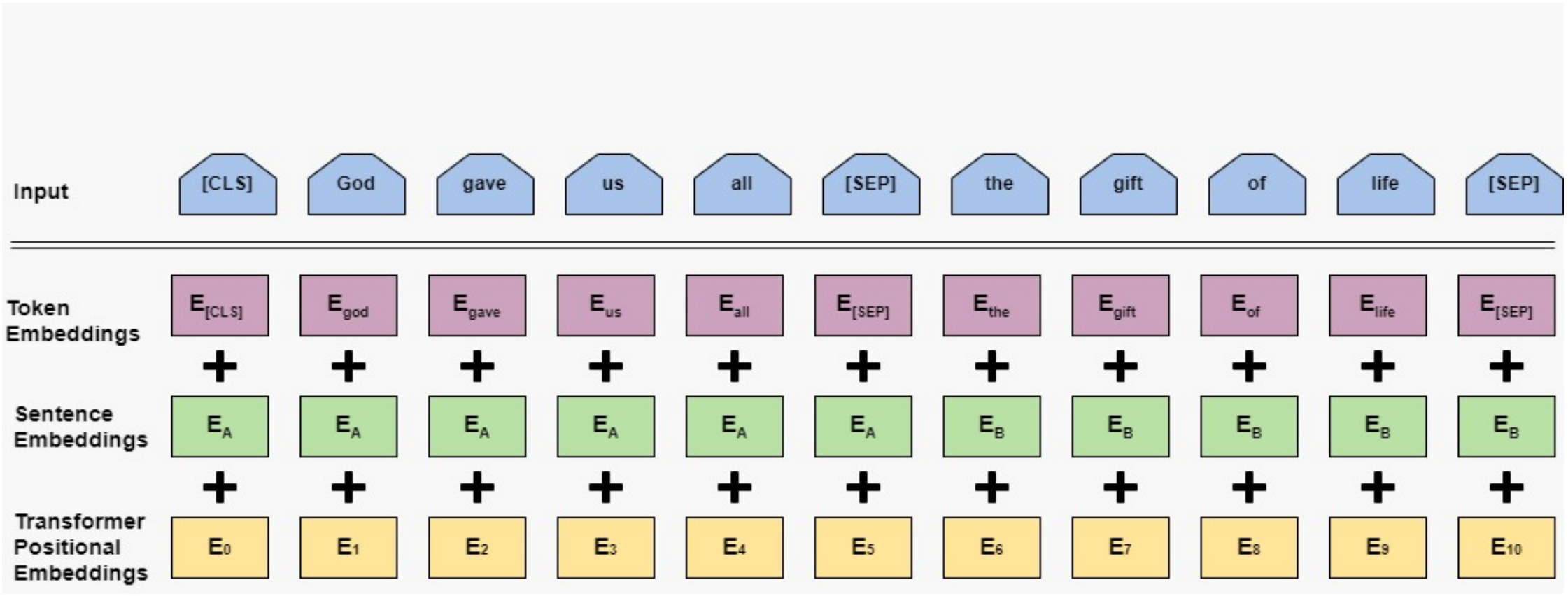

*Figure 4: Embedding in BERT*

Further, the sentences are padded, and the input is sent to the BERT encoder. After extracting the features, this step sends the output to the fully connected neural networks (shown in Fig. 5) for classification.

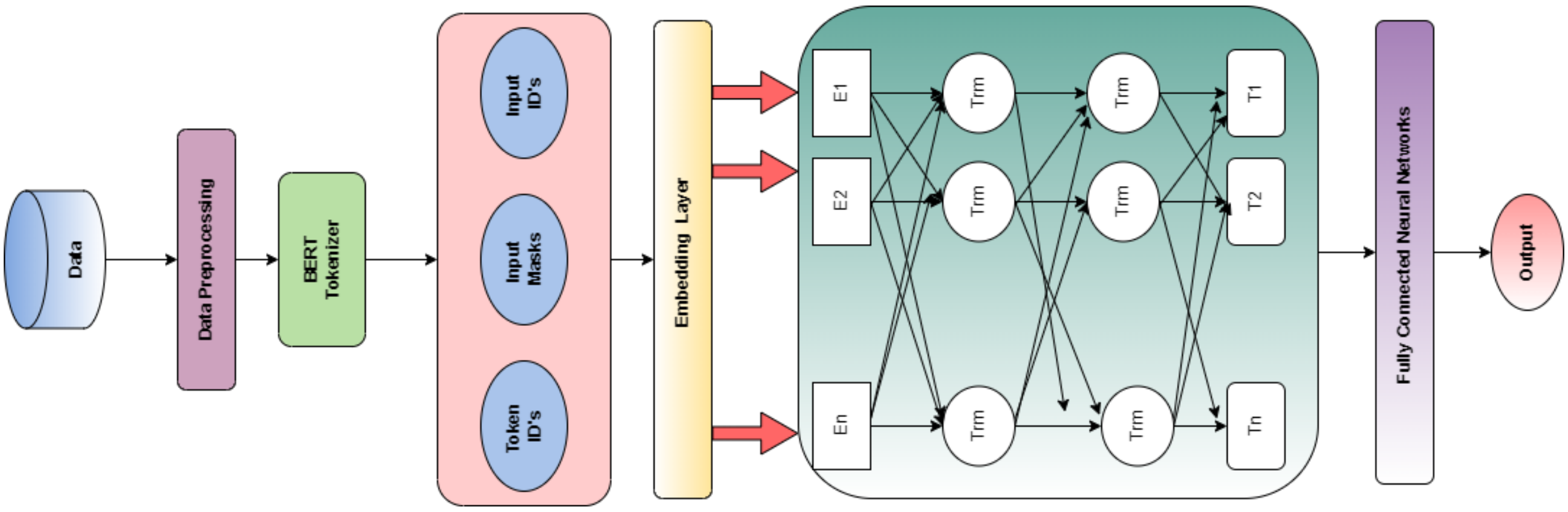

*Figure 5: BERT Model Workflow*

## 4. Results

The obtained results were tested against Precision, Recall and F1-Scores as the performance metrics, where

$$Precision = \frac{True\ Positive}{True\ Positive + False\ Positive}$$

$$Recall = \frac{True\ Positive}{True\ Positive + False\ Negative}$$

and

$$F1 - Score = 2\ X\ \frac{Precision\ X\ Recall}{Precision + Recall}$$

Traditional Machine Learning Models (Naïve Bayes, Logistic Regression, Support Vector Machine (SVM)) were trained on the proposed new dataset, and the following results were obtained.
The best performance was obtained using SVM, as seen in Table 5.

| Model | Label | Classification Report | | | Confusion Matrix | | |
|---|---|---|---|---|---|---|---|
| | | Precision | Recall | F1-Score | -1 | 0 | 1 |
| Naïve Bayes | -1 | 0.59 | 0.70 | 0.64 | 747 | 264 | 61 |
| | 0 | 0.60 | 0.69 | 0.65 | 277 | 687 | 26 |
| | 1 | 0.76 | 0.40 | 0.53 | 238 | 185 | 283 |
| Logistic Regression | -1 | 0.63 | 0.68 | 0.65 | 731 | 222 | 119 |
| | 0 | 0.66 | 0.67 | 0.66 | 260 | 662 | 68 |
| | 1 | 0.69 | 0.58 | 0.63 | 178 | 121 | 407 |
| SVM | -1 | 0.63 | 0.69 | 0.66 | 744 | 220 | 108 |
| | 0 | 0.66 | 0.67 | 0.66 | 259 | 663 | 68 |
| | 1 | 0.69 | 0.57 | 0.62 | 186 | 121 | 399 |

Table 5: Performance Metrics with Traditional ML Models

We also used Transfer Learning to fine-tune pre-trained models such as BERT to obtain the desired results shown below. It is noticeable from Table 6 that the results have improved a bit but still have scope for improvement.

| Model | Label | Classification Report | | | Confusion Matrix | | |
|---|---|---|---|---|---|---|---|
| | | Precision | Recall | F1-Score | -1 | 0 | 1 |
| BERT (3 Classes) | -1 | 0.69 | 0.69 | 0.69 | 1555 | 288 | 405 |
| | 0 | 0.66 | 0.72 | 0.69 | 260 | 1199 | 200 |
| | 1 | 0.73 | 0.68 | 0.71 | 436 | 329 | 1645 |

Table 6: Performance Metrics for BERT with 3 Classes

Inferring from the above confusion matrix, we observed that the model was getting confused because of the neutral statements, so we decided to train the model again without neutral statements. The results obtained are as follows. Upon training the model without neutral statements, it is observed that the results saw a significant improvement. This is shown in Table 7.

| Model | Label | Classification Report | | | Confusion Matrix | |
|---|---|---|---|---|---|---|
| | | Precision | Recall | F1-Score | -1 | 0 |
| BERT (3 Classes) | 0 | 0.80 | 0.85 | 0.83 | 591 | 105 |
| | 1 | 0.89 | 0.85 | 0.87 | 145 | 822 |

Table 7: Performance Metrics for BERT with 2 Classes

## 5. Discussion

From the confusion matrix in table 6, it is observed that the model is not able to differentiate between non-hope and neutral statements (evident as the model classified 405 neutral statements as non-hope and 436 non-hope statements as neutral). Further, which motivated us to train the model only with Hope and non-hope labels which produced satisfactory results (shown in table 7).

Furthermore, the presence of the exact words in sentences labelled as Hope and Non-Hope, the model is not able to classify them correctly. For example, text data like "If you say all lives matter "can be classified into Hope and non-hope speech. Since the word "All lives matter" gives Hope, the presence of "If you say" in the sentence makes it difficult even for annotators to classify the sentence into an appropriate label. Similarly, text data like "She got placed in a good company being a woman" can be interpreted as both Hope and non-hope because the text "being a woman" adds discrimination to the sentence. In such cases, it is difficult for the model to perform classification. The original dataset contained many contractions like BLM, msm, libs which means Black lives matter, main stream media, liberals respectively. The model identifies these contracted words as different from their expanded counterpart. Upon revisiting the dataset, these contractions were expanded. It is imperative to note that although the accuracy obtained with the previous dataset was more than that obtained with the newly modified dataset, the old dataset was faulty. Many sentences were repeated, and at one instance, labelled as Hope Speech and at another as Non-Hope Speech. Hence, the results obtained on the new dataset cannot and should not be compared with the previous dataset due to its various shortcomings.

## 6. Conclusion and Future Work

Hence, we propose a revised dataset for the problem statement "Hope speech detection for equality, diversity and inclusion", with the introduction of a new third label to properly classify the statements and reduce misclassifications. Best results were obtained through implementing BERT i.e., Precision as 0.6973, recall as 0.6966, F1-score as 0.6966. On removing the statements with the neutral label, we obtained results from BERT: Precision as 0.845, Recall as 0.85, and F1-score as 0.85.

In the future, the scope of this approach can be expanded by including multiple languages and considering sarcasm. Further, this solution might help take content moderation on social media platforms one step further by only highlighting hopeful content.